\documentclass[lettersize,hidelinks,journal]{IEEEtran}
\usepackage{amsmath,amsfonts}
\usepackage{algorithmic}
\usepackage{algorithm}
\usepackage{array}
\usepackage[caption=false,font=normalsize,labelfont=sf,textfont=sf]{subfig}
\usepackage{textcomp}
\usepackage{stfloats}
\usepackage{url}
\usepackage{verbatim}
\usepackage{graphicx}
\usepackage{cite}
\usepackage{booktabs}
\usepackage{multirow}
\usepackage{multicol}
\usepackage{orcidlink}
\usepackage{arydshln}
\usepackage{threeparttable}
\usepackage{adjustbox}

\usepackage{pgfplots, tikz}
\usetikzlibrary{patterns}
\pgfplotsset{compat=1.18}

\usepackage{pifont}

\begin{document}

\title{Tailored Design of Audio-Visual Speech Recognition Models using Branchformers}

\author{David Gimeno-Gómez\orcidlink{0000-0002-7375-9515}, Carlos-D. Martínez-Hinarejos\orcidlink{0000-0002-6139-2891} 
\thanks{Pattern Recognition and Human Language Technology research center, Universitat Politècnica de València, Spain (e-mail: dagigo1@dsic.upv.es)}
\thanks{This version of the article has been accepted for publication in Computer Speech \& Language, after peer review. The final published version is available at: {\color{blue}\url{https://doi.org/10.1016/j.csl.2025.101811.}}}}

\markboth{Journal of \LaTeX\ Class Files,~Vol.~XX, No.~X, XXXX~XXXX}%
{Shell \MakeLowercase{\textit{et al.}}: A Sample Article Using IEEEtran.cls for IEEE Journals}

\IEEEpubid{0000--0000/00\$00.00~\copyright~2021 IEEE}

\maketitle

\begin{abstract}
Recent advances in Audio-Visual Speech Recognition (AVSR) have led to unprecedented achievements in the field, improving the robustness of this type of system in adverse, noisy environments. In most cases, this task has been addressed through the design of models composed of two independent encoders, each dedicated to a specific modality. However, while recent works have explored unified audio-visual encoders, determining the optimal cross-modal architecture remains an ongoing challenge. Furthermore, such approaches often rely on models comprising vast amounts of parameters and high computational cost training processes. In this paper, we aim to bridge this research gap by introducing a novel audio-visual framework. Our proposed method constitutes, to the best of our knowledge, the first attempt to harness the flexibility and interpretability offered by encoder architectures, such as the Branchformer, in the design of parameter-efficient AVSR systems. To be more precise, the proposed framework consists of two steps: first, estimating audio- and video-only systems, and then designing a tailored audio-visual unified encoder based on the layer-level branch scores provided by the modality-specific models. Extensive experiments on English and Spanish AVSR benchmarks covering multiple data conditions and scenarios demonstrated the effectiveness of our proposed method. Even when trained on a moderate scale of data, our models achieve competitive word error rates (WER) of approximately 2.5\% for English and surpass existing approaches for Spanish, establishing a new benchmark with an average WER of around 9.1\%. These results reflect how our tailored AVSR system is able to reach state-of-the-art recognition rates while significantly reducing the model complexity w.r.t. the prevalent approach in the field. Code and pre-trained models are available at \url{https://github.com/david-gimeno/tailored-avsr}.
\end{abstract}

\begin{IEEEkeywords}
Audio-visual speech recognition, branchformer, architecture design, unified audio-visual encoder, interpretability, parameter efficiency.
\end{IEEEkeywords}

\section{Introduction}
\IEEEPARstart{A}{utomatic} Speech Recognition (ASR) is a widely studied task in the field of speech processing \cite{06gales2008application}. Albeit current ASR systems are nowadays capable of understanding spoken language with great quality \cite{radford2023robust,baevski2020wav2vec}, their performance can be considerably affected in adverse scenarios, such as a noisy environment \cite{juang1991adverse}. This challenge has led to the emergence of Visual Speech Recognition (VSR) technologies as an interesting avenue of research \cite{fernandez2018survey,ma2022visual,chang2024allyouneed}. Nonetheless, inspired by our speech perception process \cite{mcgurk1976hearing,besle2004bimodal,stacey2020audio}, researchers have increasingly turned their focus towards Audio-Visual Speech Recognition (AVSR) as a promising solution, demonstrating how the incorporation of visual cues when interpreting speech improves the performance of this type of systems in challenging, noisy scenarios \cite{afouras2018deep,ma2021conformers,shi2022learning,burchi2023audio,ma2023auto,fernandezlopez2024msrs}. While multiple methods have been investigated to effectively integrate both modalities \cite{potamianos2003advances,sterpu2018fusion,wei2020attentive,wu2021cocktail}, the prevalent AVSR architecture nowadays is based on early fusion strategies, either combining the speech features extracted by two independent encoders \cite{afouras2018deep,ma2021conformers,ma2023auto,burchi2023audio}, each dedicated to a specific modality, or adopting the recent trend of unified multi-modal encoders \cite{shi2022learning,zhu2023vatlm,li2023unified}. 

\IEEEpubidadjcol
However, although these approaches represent the current state of the art, determining the optimal cross-modal architectures remains an ongoing challenge. A significant limitation of existing studies lies not only in their reliance on models with vast amounts of parameters and extensive pretraining processes, but also in their focus on how audio-visual features should be fused to enhance the performance of a fixed encoder backbone. Therefore, it is not clear how encoder architectures can be adapted to the different nature and intrinsic characteristics of each modality. The Branchformer encoder \cite{peng2022branchformer,kim2023ebranchformer} seemed to represent one step forward in this regard. However, apart from being proposed solely in the context of auditory-based ASR, the authors did not exploit the potential of the flexibility and interpretability offered by this architecture for the design of tailored and parameter-efficient systems. These were the reasons that motivated our work, whose key contributions are:

\begin{itemize}
    \item We propose a novel framework to design tailored, unified audio-visual encoders capable of reaching state-of-the-art performance, while reducing the number of model parameters by nearly half, e.g., from 103.5M to 59.3M in the model trained for English, when compared to the conventional and prevalent AVSR architecture.

    \item To the best of our knowledge, our work is one of the first attempts to harness the flexibility and interpretability offered by encoder architectures, such as the Branchformer, in the design of parameter-efficient AVSR systems. Specifically, we leverage these properties to identify the context dependencies -- whether short- or long-term -- that are more relevant for processing each modality. This analysis directly informs the design of a tailored AVSR system, optimized to handle the distinct characteristics of audio and visual inputs effectively.

    \item Through extensive experiments on two English (LRS2-BBC \cite{afouras2018deep} and LRS3-TED \cite{afouras2018lrs3}) and four Spanish (MuAViC \cite{anwar23muavic}, CMU-MOSEAS \cite{zadeh2020moseas}, LIP-RTVE \cite{lrec2022liprtve}, and VLRF \cite{fernandez2017towards}) AVSR benchmarks, we demonstrated the effectiveness of our proposed audio-visual framework across diverse data conditions and scenarios. Even with a moderate scale of training data and a significantly reduced number of model parameters, our models remarkably achieve competitive word error rates (WER) of approximately 2.5\%, on par with state-of-the-art methods \cite{ma2021conformers,burchi2023audio} under similar conditions. For Spanish datasets, the framework surpasses existing approaches, achieving an average WER around 9.1\%, thereby establishing a new benchmark for this language.

    \item Our analysis provides initial insights into the architectural differences, beyond the frontends, in processing speech features across modalities. Acoustic cues rely on both short- and long-term relationships, whereas the video-only system heavily depends on global-context cues, likely reflecting the ambiguity and limited information in lipreading. Additionally, our adaptive audio-visual fusion module underscored the dominant role of acoustic cues, with a relevance of approximately 70\% --- a finding that aligns with studies indicating that only 30\% of speech information is visible \cite{duchnowski2000development}.

    \item We investigate the impact of noise distortions during both training and inference, as well as scenarios where the acoustic signal is completely absent. Consistent with \cite{lin2024uncovering}, our findings highlight the limitations of current approaches in effectively exploiting the complementary information provided by visual cues, particularly in our Spanish-language experiments.

\end{itemize}

\section{Related Work}

\noindent\textbf{State of the Art in AVSR.} Thanks to the collection of large-scale datasets \cite{afouras2018lrs3,anwar23muavic} and the use of powerful attention-based mechanisms \cite{vaswani2017attention}, unprecedented results have been achieved in the context of AVSR \cite{shi2022learning,burchi2023audio,ma2023auto,chang2024allyouneed}. While most of these studies are predominantly evaluated on English --- with performances between 0.8-0.9\%, 1.0-1.5\%, and 12-20\% WER for AVSR, ASR, and VSR tasks, respectively \cite{ma2023auto,chang2024allyouneed} --- the performance for other languages, such as Spanish, remains comparably distant. Though recent works are considering more languages in their studies \cite{ma2022visual,kim2023lip,anwar23muavic,hong2023intuitive}, Spanish AVSR performance is notably lower with an average WER of approximately 16\%, 15\%, 50\% WER for AVSR, ASR, and VSR tasks, respectively. In terms of model architecture, the hybrid CTC/Attention framework \cite{watanabe2017ctcattention} is the most prevalent decoding paradigm adopted in the field. In our research, we also adopt this architecture, aiming to assess the effectiveness of our proposed audio-visual encoder more compellingly.

\noindent\textbf{Parameter-Efficient ASR Models.} Several works have focused their efforts on building parameter-efficient ASR models through diverse methods, including distillation \cite{li2018distillation}, quantization \cite{nguyen2020quantization,sainath2020streaming}, pruning \cite{lai2021parp,ding2022lottery}, and parameter-sharing techniques \cite{wei2023sim}. Compatible with structured sparsity, Ding et al. \cite{ding2022lottery} investigated the search of optimal sparse subnetworks within audio-only ASR architectures, not only without experiencing a significant drop in performance but also achieving robustness under noisy conditions. Although limited, similar methods were also applied for VSR, highlighting Fernández-López et al. \cite{fernandez2023sparsevsr}, which explored multiple pruning strategies, also demonstrating the potential robustness of these sparse models to noise compared to their dense counterparts. For AVSR, the increasing computational demands have led to innovations such as mask optimization training procedures aimed at improving scalability \cite{fernandezlopez2024msrs}. Of particular note, however, is the work by Burchi and Timofte \cite{burchi2023audio}, who introduced a non-autoregressive CTC-based AVSR architecture. Their approach achieved competitive results, while reducing model complexity through an efficient attention mechanism. However, unlike our research, their proposed model offered no architectural innovation in the fusion of audio-visual speech cues.

\noindent\textbf{Interpreting ASR.} Multiple studies demonstrated that as we progress through the encoder layers, low-level patterns like speaker characteristics and noise are gradually disregarded, leading to a more standardized representation of high-level semantics \cite{mohamed2012understanding,li2020hear}. The primary focus of other works has been to interpret the underlying phonetic representations captured by these deep neural networks. This was achieved either through phonetic-based regularization techniques during training \cite{tan2015stimulated,chunyang2018regularization} or the incorporation of additional modules to provide more explainable speech representations, with the goal of identifying the most relevant features for the model \cite{ravanelli2018interpretable,agrawal2020relevance}. While these studies focused on improving the interpretation of ASR systems, the Branchformer encoder \cite{peng2022branchformer} represents one of the first attempts in addressing the problem from the perspective of model architecture design. As described in Subsection \ref{sec:architecture}, this encoder allows researchers to interpret whether the global or local dependencies are more relevant at each encoder layer when processing speech. However, despite conducting several interesting studies, the authors did not exploit the potential of this novel approach towards the design of optimal and parameter-efficient ASR architectures as we propose in this paper. In addition, unlike our work, all these studies were focused on auditory-only ASR without considering the audio-visual integration and how both modalities differentiate and complement each other.

\noindent\textbf{Cross-Modal Unified Encoders for AVSR.} Shi et al. \cite{shi2022learning} introduced AV-HuBERT, a self-supervised learning framework for AVSR capable of reaching state-of-the-art results on the LRS3-TED benchmark \cite{afouras2018lrs3}. The audio- and video-based cues are first processed by modality-independent frontends. These frontend features are then fused and fed into a shared backbone Transformer estimated with a frame-level masked cluster prediction task. Zhu et al. \cite{zhu2023vatlm} presented V{\footnotesize AT}LM, an extension of the AV-HuBERT framework that not only integrates audio and visual cues, but it also incorporates the corresponding text transcription as an additional input to the unified cross-modal encoder backbone. Both architectures presented a base and large variant comprising around 105M and 330M, respectively. However, the substantial performance gap between the base and large models in both works reflects how the effectiveness of these architectures highly depends on the number of parameters that they comprise. Similar findings are found for the cross-lingual XLAVS-R model \cite{han2024xlavs}, whose architecture is mainly based on AV-HuBERT. Conversely, Li et al. \cite{li2023unified} 
compared multiple methods to fuse the audio and visual cues using a supervised AVSR architecture without relying on large-scale pretraining processes. Their experiments showed superior performance when concatenating both modalities along the temporal dimension, thus offering a robust approach to noisy scenarios that does not require forced alignment despite the modality mismatch in terms of sampling rate. 

\noindent\textbf{Our present work.} However, these studies were limited to how the audio-visual features should be fused to maximize the performance of a fixed unified encoder, which was not modified or adapted to the different nature and intrinsic characteristics of each modality. Unlike prior research, our proposed method focused on the tailored design of the optimal cross-modal encoder architecture for parameter-efficiency AVSR by leveraging the interpretability and flexibility of the Branchformer encoder \cite{peng2022branchformer}. Furthermore, we show the effectiveness of our proposed method through extensive experiments in two different languages and a wide range of data conditions, even when estimated with a moderate amount of training data.

\section{Method} \label{sec:method}

For clarity, we first describe the conventional and prevalent end-to-end architecture for AVSR in Subsection \ref{sec:architecture}, as background, to subsequently introduce our proposed tailored and unified audio-visual encoder in Subsection \ref{sec:tailored}.

\subsection{Model Architecture} \label{sec:architecture}

\noindent\textbf{Audio Frontend.} The audio frontend transforms the raw audio waveforms into 80 mel-spectrogram coefficients by applying a short-time Fourier transform with a 20ms window and a 10ms step size. Then, a two-layer convolutional module is not only in charge of extracting more powerful local and temporal acoustic features, but also of down-sampling the pre-processed audio sequence to a quarter of its length. This down-sampling addresses the mismatch in sample rates between audio and visual cues, which typically correspond to 100 fps and 25 fps, respectively, as it is the case with the datasets in our case study. Consequently, audio and visual cues are temporally aligned. The final acoustic representation sequence consists of 256-dimensional embeddings.

\noindent\textbf{Visual Frontend.} Consistent with prior works \cite{ma2022visual,ma2023auto}, spatio-temporal relationships are captured by using a 3D convolutional layer with a kernel of 7$\times$7 pixels and a receptive field of 5 frames. Once these visual features are flattened along the time dimension, a 2D ResNet-18 \cite{he2016resnet} is responsible for extracting local visual patterns, resulting in a 512-dimensional feature representation.

\noindent\textbf{Branchformer Encoder.} Both the acoustic and visual feature sequences are projected to a 256-dimensional space and further injected
with a relative positional encoding \cite{dai2019transformerxl}. Then, these enhanced frontend representations are each passed through a different modality-specific encoder inspired by the Branchformer architecture \cite{peng2022branchformer} and its enhanced variant, E-Branchformer \cite{kim2023ebranchformer}. This enhanced variant not only explored more advanced convolution-based merging techniques, but also incorporated feed-forward modules in a macaron-style configuration. Each of the 12 layers in our encoder consists primarily of two parallel branches:

\begin{itemize}
    \item \textbf{Self-Attention Branch.} This branch is responsible for modeling the global long-term dependencies from speech cues using a multi-headed self-attention mechanism \cite{vaswani2017attention}. Concretely, we set to 4 and 64 the number and size of the attention heads comprising the module, respectively.
    \item \textbf{cgMLP Branch.} A Multi-Layer Perceptron with Convolutional Gating (cgMLP) module \cite{sakuma2022mlpbased} is designed to capture the local context relationships. Specifically, the input sequence is first projected to an up-sampling inner dimension of 2048. Then, the sequence is split to process part of it using a 1D depth-wise convolutional layer with a kernel size of 31 to subsequently apply a linear gating mechanism. Finally, a linear layer projects the obtained representation to its original dimension of 256. 
\end{itemize}

\begin{figure}[!htbp]
\centering
\includegraphics[width=0.85\columnwidth]{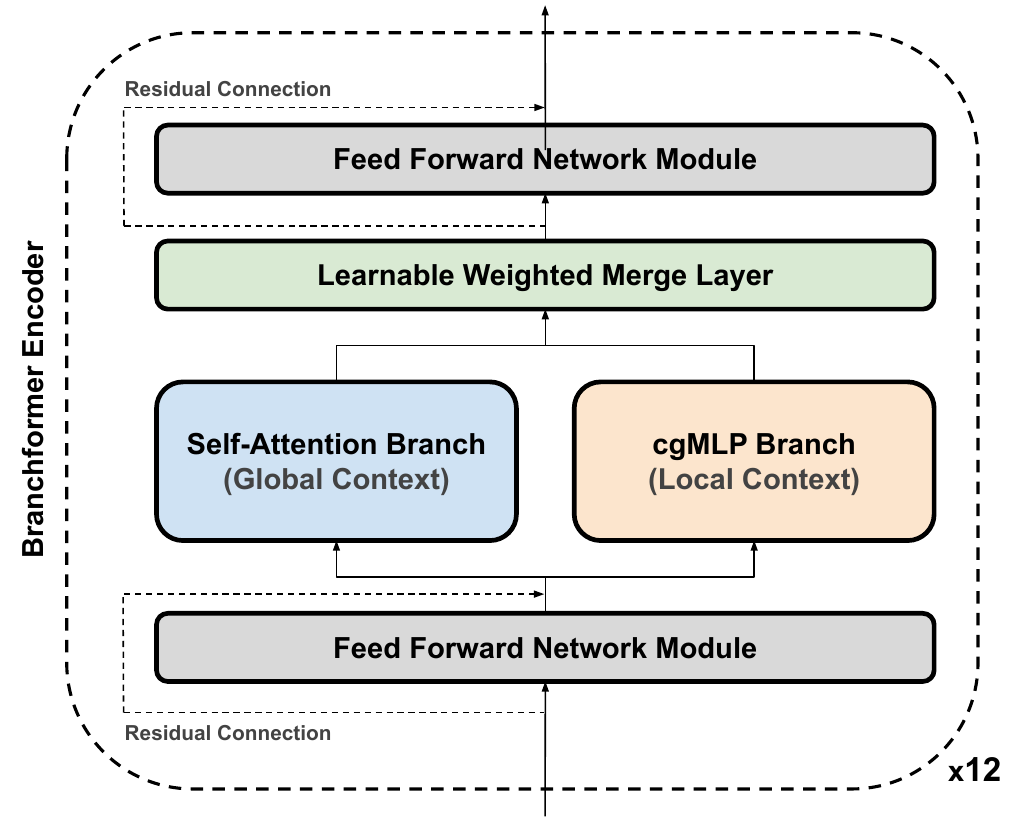}
\caption{Architecture design of an encoder Branchformer layer \cite{peng2022branchformer}.}
\label{fig:branchformer}
\end{figure}

As shown in the scheme depicted in Figure \ref{fig:branchformer}, the encoder layers also incorporate two feed-forward modules with an up-sampling inner dimension of 2048 set in a macaron-style manner alongside their corresponding residual connections with a scale factor of $1/2$. In all cases, the Swish activation function \cite{swish2017prajit} is used, layer normalization \cite{ba2016layer} precedes each of the described modules, while a final dropout \cite{srivastava2014dropout} of 0.1 is applied over their output. Finally, the latent representations provided by both branches are merged via a learnable weighted method that, as described below (and reflected in Figure \ref{fig:heatmaps}), allowed us to interpret which branch was more relevant at each encoder layer for the audio- and video-only case studies.

\noindent\textbf{Learnable Weighted Merge Layer.} This module is in charge of combining the output of both branches and dynamically generating the weights representing the relevance of each branch at each encoder layer through an attention-based pooling projection process. Let $\boldsymbol{Z} \in \mathbb{R}^{T \times d}$ be the $d$-dimensional audio or video sequence of length $T$ processed by one  of the encoder branches and $\boldsymbol{w} \in \mathbb{R}^{d}$ be a learnable parameter vector. The attention weights $\boldsymbol{\alpha} \in \mathbb{R}^{T}$ for each vector $\boldsymbol{z}_{i} \in \mathbb{R}^{d}, 1 \leq i \leq T$ composing $\boldsymbol{Z}$ are computed as:

\begin{equation} \label{eq:attention-pooling}
    \alpha_{i} = \frac{\exp\left(\boldsymbol{w}^{T}\boldsymbol{z}_{i} / \sqrt{d}\right)}{\Sigma^{T}_{j=1}\exp\left(\boldsymbol{w}^{T}\boldsymbol{z}_{j} / \sqrt{d}\right)},
\end{equation}

\noindent representing the scaled dot product between $\boldsymbol{Z}$ and $\boldsymbol{w}$ followed by a softmax normalization. Once the attention scores $\boldsymbol{\alpha}$ were computed, we transform the sequence $\boldsymbol{Z}$ into the output $\boldsymbol{a} \in \mathbb{R}^{d}$ via a weighted sum: $\boldsymbol{a} = \Sigma^{T}_{i=1} \alpha_{i}\boldsymbol{z}_{i}$. Therefore, based on this procedure, we could transform the output sequences from the self-attention and cgMLP branches into the attention-enriched representations denoted by $\boldsymbol{a_{att}}$ and $\boldsymbol{a_{mlp}}$, respectively. Then, these two vectors are projected to two scalars and normalized using softmax to obtain the branch relative weights as follows: 

\begin{equation} \label{eq:weight-balance}
    w_{attn}, w_{mlp} = softmax\left(\boldsymbol{W}_{attn}\boldsymbol{a}_{attn}, \boldsymbol{W}_{mlp}\boldsymbol{a}_{mlp}\right),
\end{equation}

\noindent where $\boldsymbol{W}_{att}, \boldsymbol{W}_{mlp} \in \mathbb{R}^{1 \times d}$ are linear transforms. Finally, the encoder provides a merged output $\boldsymbol{X} \in \mathbb{R}^{T \times d}$ capturing both global and local speech patterns computed as defined in:

\begin{equation} \label{eq:merged-output}
    \boldsymbol{X} = w_{att}\boldsymbol{Z}_{att} + w_{mlp}\boldsymbol{Z}_{mlp},
\end{equation}

\noindent where $\boldsymbol{Z}_{att}$ and $\boldsymbol{Z}_{mlp}$ refer to the output latent representations provided by the self-attention and cgMLP encoder branches, respectively. In this work, $d$ corresponds to 256 dimensions.

\noindent\textbf{Adaptive Audio-Visual Fusion.} We present an adaptive audio-visual fusion module based on the learnable weighted merge layer described above. Let $\boldsymbol{X}_{audio}$ and $\boldsymbol{X}_{video}$ be the audio and video sequences encoded by the corresponding modality-specific encoder, respectively, the audio-visual speech representation is defined such that:

\begin{equation} \label{eq:audiovisual-merge}
    \boldsymbol{X} = PoswiseFFN\left(w_{audio}\boldsymbol{X}_{audio} + w_{video}\boldsymbol{X}_{video}\right),
\end{equation}

\noindent where $w_{audio}$ and $w_{video}$ are the modality relative weights resulting from an attention-based pooling projection process, and $PoswiseFFN$ represents a 2-layer position-wise feed-forward module with an inner and output dimension of 2048 and 256, a Swish activation layer, and a dropout rate of 0.1. 

\noindent\textbf{CTC/Attention Decoder.} This state-of-the-art hybrid architecture \cite{watanabe2017ctcattention} combines both the Markov assumptions of CTC \cite{graves2006connectionist} (properties in harmony with the speech nature) and the flexibility of the non-sequential alignments offered by the attention-based decoder \cite{vaswani2017attention}. As Figure \ref{fig:tailored} depicts, this module is composed of a linear projection as the CTC-based decoding branch and a 6-layer Transformer decoder of 4 attention heads of 64 dimensions attached to the Cross Entropy (CE) criterion. As proposed by \cite{watanabe2017ctcattention}, its loss function is defined as follows:

\begin{equation} \label{eq:ctc-attention-loss}
    \mathcal{L} = \alpha \log p_{ctc}(\boldsymbol{Y}|\boldsymbol{X}) + (1 - \alpha) \log p_{attn}(\boldsymbol{Y}|\boldsymbol{X}),
\end{equation}

\noindent where $p_{ctc}$ and $p_{attn}$ denote the CTC and Attention posteriors, respectively. In both terms, $\boldsymbol{X}$ and $\boldsymbol{Y}$ refer to the speech latent representation processed by the encoder and its corresponding character-level target transcription, respectively.  The $\alpha$ weight balances the relative influence of each decoding branch.

\noindent\textbf{Language Model.} In this work, we use the character-level Language Models (LM) publicly released by Ma et al. \cite{ma2022visual} for Spanish and English. These models, defined as 16-layer Transformer encoders \cite{vaswani2017attention}, were estimated with text from different sources comprising more than 150M characters each.

\noindent\textbf{Inference.} In the CTC/Attention architecture inference process, the Attention-based decoder primarily determines when the beam search decoding concludes by predicting the end-of-sentence token, and thus both the CTC-based decoding branch and the LM serve as additional scoring factors, as reflected by

\begin{equation} \label{eq:decoding}
    S = \lambda S_{ctc} + (1 - \lambda)S_{attn} + \beta S_{lm},
\end{equation}

\noindent where $S_{ctc}$ and $S_{attn}$ are the scores of the CTC- and the Attention-based decoder, respectively, $\lambda$ is their corresponding relative weight, and $\beta$ and $S_{lm}$ refer to the LM decoding influence weight and the LM score, respectively. Readers are referred to Watanabe et al. \cite{watanabe2017ctcattention} for a more comprehensive description of this inference process.

\subsection{The Tailored Audio-Visual Branchformer} \label{sec:tailored}

The model architecture previously described, composed of two independent modality-specific encoders, corresponds to the most prevalent approach adopted in the context of AVSR \cite{afouras2018deep,ma2021conformers,prajwal2022sub,ma2023auto,burchi2023audio}. While recent works have explored unified encoder architectures \cite{shi2022learning,zhu2023vatlm,li2023unified}, determining the optimal cross-modal architecture remains an ongoing challenge. In this paper, unlike prior works, we introduce a novel framework that, to the best of our knowledge, constitutes the first attempt to harness the flexibility and interpretability offered by encoder architectures, such as the Branchformer \cite{peng2022branchformer}, in the design of parameter-efficient AVSR systems. The proposed method comprises multiple steps.

\noindent\textbf{Training Modality-Specific Models.} We first estimated models for audio-only and video-only inputs separately, each utilizing its corresponding modality-specific Branchformer encoder: one dedicated to the processing of audio speech cues ($E_{audio}$) and the other to visual speech cues ($E_{video}$).

\begin{figure*}[!t]
\centering
\subfloat[]{\includegraphics[width=2.85in]{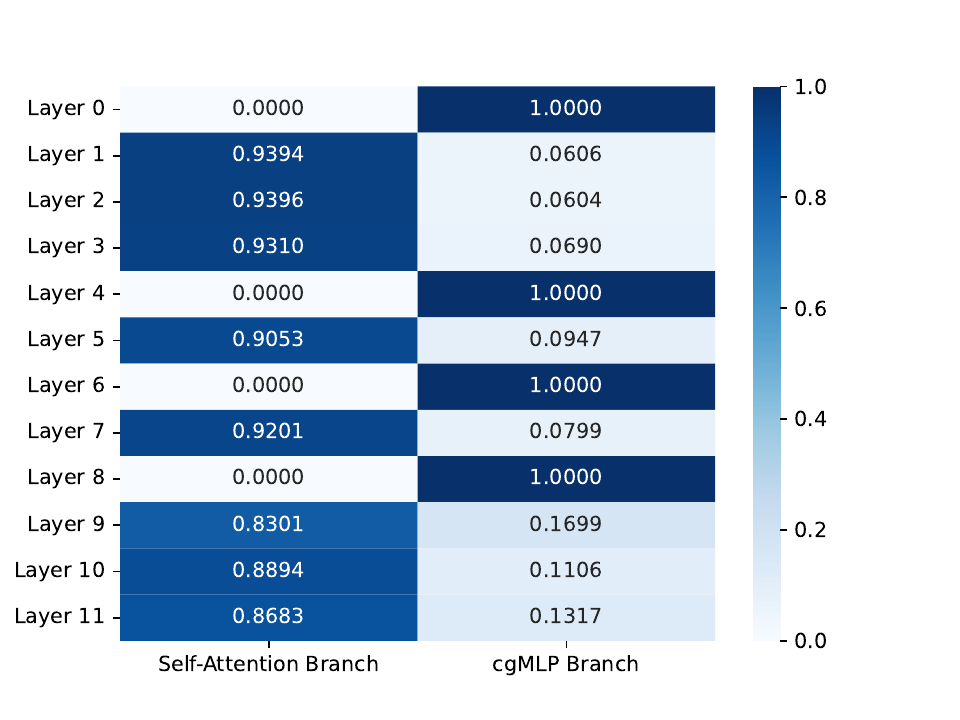}%
\label{asr-heatmap}}
\hfil
\subfloat[]{\includegraphics[width=2.85in]{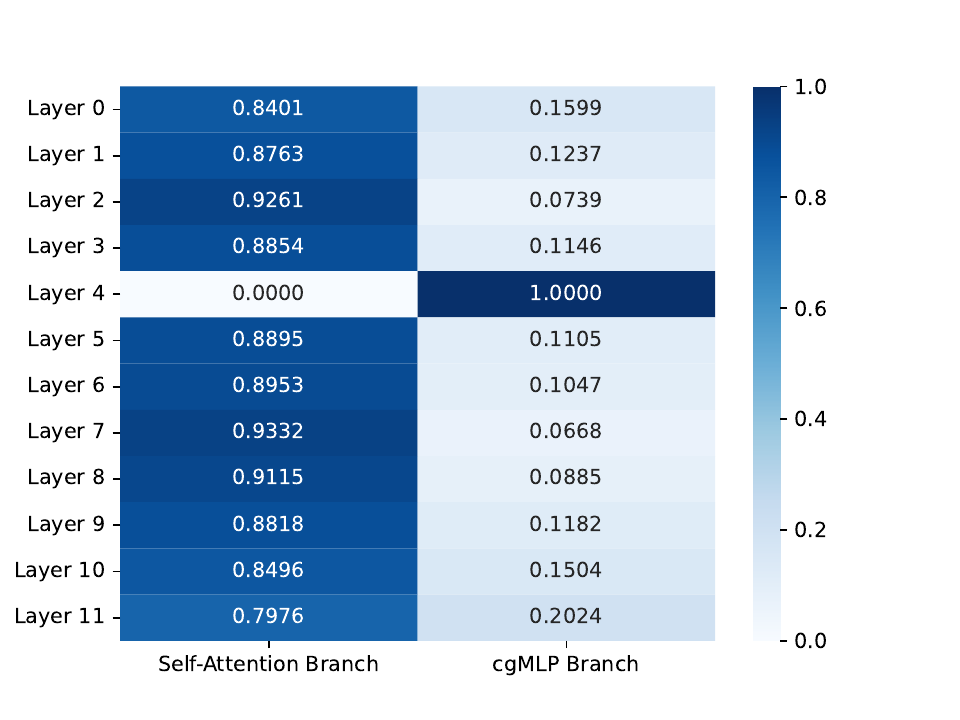}%
\label{vsr-heatmap}}
\caption{Visualization of the average Branchformer weights on the LRS2 validation data set. (a) Acoustic Speech Recognition. (b) Visual Speech Recognition.}
\label{fig:heatmaps}
\end{figure*}

\noindent\textbf{Analyzing Branchformer Encoder Layers.} As described in Subsection \ref{sec:architecture}, the Branchformer encoder allowed us to analyze which branch was more relevant at each encoder layer for each of our estimated modality-specific models. Figure \ref{fig:heatmaps} reflects how the encoder adopted a different architecture depending on the addressed modality for the LRS2-BBC dataset. The increased reliance on global context relationships in the video-only scenario, compared to the audio-only setting, could be attributed to the limited information provided by lip movements when interpreting speech \cite{duchnowski2000development}. Section \ref{sec:results} discusses the patterns observed across the datasets explored in our work.

\begin{figure*}[!htbp]
\centering
\includegraphics[width=\textwidth]{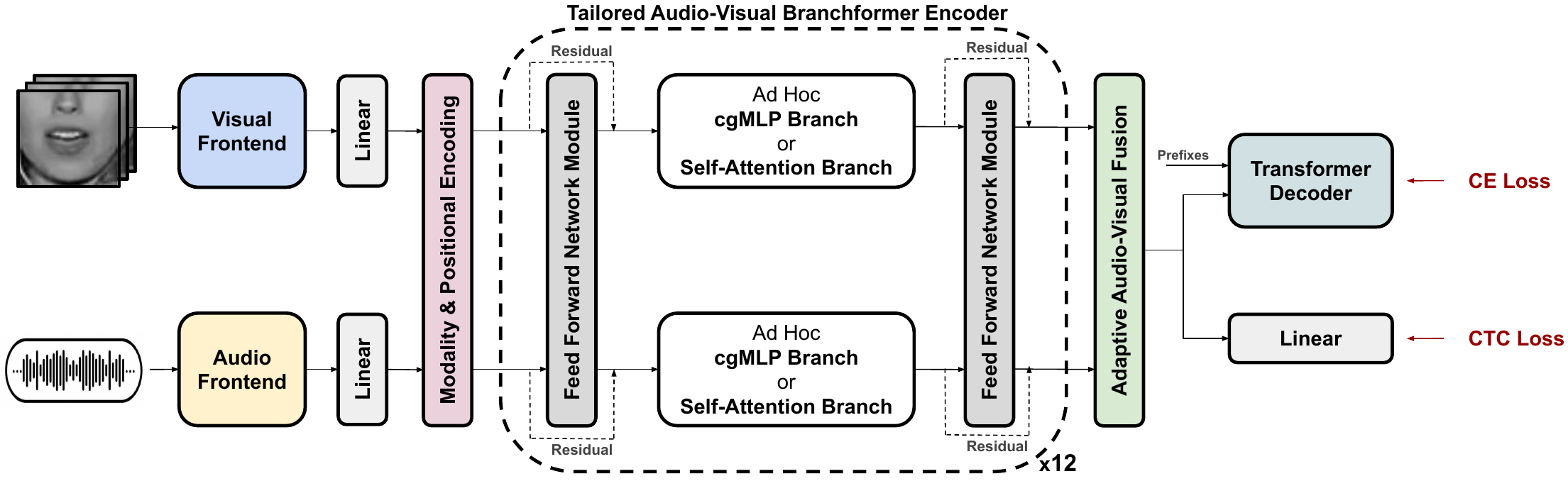}
\caption{Our proposed method involves preprocessing visual and audio cues, conditioning the resulting temporal-aligned speech features with positional and modality embeddings, and then modeling them using a tailored encoder, whose architecture design is based on the interpretable weights of models previously estimated for audio- and video-only settings, to perform the final speech interpretation in an autoregressive manner through the hybrid CTC/Attention paradigm. Both feed forward networks share parameters across modalities. CE and CTC refer to Cross Entropy and Connectionist Temporal Classification, respectively.}
\label{fig:tailored}
\end{figure*}

\noindent\textbf{Designing a Tailored AVSR Model.} We also observed that, in most cases and regardless of the modality, one of the relative branch weights of the modality-specific models tended to approach saturation, with values consistently close to 1. Our hypothesis was that, due to this saturation, the branch associated with the lower impact could be removed without a significant decrease in model performance. Based on this analysis, we designed a tailored, unified, parameter-efficient encoder for AVSR, whose overview scheme is depicted in Figure \ref{fig:tailored}. This tailored design involves the selection of ad-hoc global- or local-context processing modules depending on the modality and the specific encoder layer level we are addressing. Specifically, our proposed tailored encoder is composed of 12 layers with two parallel branches but, in this case, each one is in charge of processing a different modality. Each layer of these branches is defined by a self-attention or cgMLP module depending on which of them was more relevant in its corresponding modality-specific model. Formally, this can be expressed as follows:

\begin{equation} \label{eq:tailored}
    \delta_{im}(x_{m}) = \begin{cases}
        cgMLP(x_{m}) &     \text{if } E_{im}^{global}(x_{m}) < E_{im}^{local}(x_{m}),\\
        SelfAttention(x_{m}) &    \text{otherwise,}
    \end{cases}
\end{equation}

\noindent where $i$ denotes the layer index, $m \in \{audio, video\}$ indicates the modality, and $x_{m}$ represents the input features for modality $m$. Therefore, $\delta_{im}$ corresponds to the context-specific processing module of the $i$-th layer of the tailored, unified audio-visual encoder. Additionally, $E_{im}^{global}(x_{m})$ and $E_{im}^{local}(x_{m})$ refer to the scores provided by our modality-specific models that indicate the relevance of global and local context at layer $i$ for modality $m$, respectively.

Unlike the original Branchformer encoder, we do not merge the encoded representation of each branch. Doing so would fundamentally alter the nature of the audio- and video-only representations, thus hindering the ad-hoc design described above by not having a base model to select one module focused on either global or local context relationships. Nonetheless, we share the two feed-forward modules with the intention of enhancing the model's cross-modal capabilities. Consequently, in order to facilitate the model to distinguish between both modalities, we introduced a modality embedding that conditions the frontend speech representations. A preliminary ablation study showed that, although no significant differences were observed, incorporating the modality embedding consistently led to slight improvements.

As supported by our extensive experiments, this tailored architecture design allowed us to reduce the number of model parameters by nearly half compared to conventional AVSR systems, while achieving state-of-the-art recognition rates across diverse benchmarks.

\section{Experiments}

The models proposed in this work were implemented using the open-source ESPNet toolkit \cite{watanabe18espnet}. Experiments were conducted on a 24-core 3.40MHz Intel i7-13700K CPU and a GeForce RTX 4090 GPU with 24GB memory.

\subsection{Datasets} \label{sec:databases}

In our work, we considered multiple AVSR benchmarks covering two different languages, as well as a wide range of scenarios and data conditions, to properly evaluate the effectiveness of our proposed audio-visual framework.

\noindent\textbf{LRS2-BBC} \cite{afouras2018deep} is a large-scale English database composed of around 224 hours collected from BBC TV programs. It consists of a pre-training set with 96,318 samples (195 hours), a training set with 45,839 samples (28 hours), a validation set with 1,082 samples (0.6 hours), and a test set with 1,243 samples (0.5 hours). It offers more than 2M running words with a vocabulary size of around 40k.

\noindent\textbf{LRS3-TED} \cite{afouras2018lrs3} is the largest publicly audio-visual English database, providing around 438 hours collected from TED talks. It consists of a `pre-train' set with 118,516 samples (407 hours), a `train-val' set with 31,982 samples (30 hours), and a test set with 1,321 samples (0.9 hours).  It offers more than 4M running words with a vocabulary size of around 50k.

\noindent\textbf{MuAViC\textsubscript{es}} \cite{anwar23muavic} is a large-scale multilingual corpus collected from TED and TEDx talks. In our work, we only considered its Spanish partition, offering around 181 hours of data. It is split into a training set with 102,035 samples (177.5 hours), a validation set with 906 samples (1.6 hours), and a test set of 1012 samples (1.7 hours). It comprises around 1.7M running words with a vocabulary size of around 58k.

\noindent\textbf{CMU-MOSEAS\textsubscript{es}} \cite{zadeh2020moseas} is a large-scale database collected from YouTube monologue videos covering 4 different languages. In our work, we only considered its Spanish partition. According to \cite{ma2022visual}, we defined a training set with 8,253 samples (15.7 hours) and a test set with 329 samples (0.6 hours). The database provides more than 150k running words with a vocabulary size of around 14k.

\noindent\textbf{LIP-RTVE} \cite{lrec2022liprtve} is a challenging Spanish database collected from TV newscast programs, providing around 13 hours of data. Its speaker-independent partition consists of a training set with 7,142 samples (9 hours), a validation set with 1638 samples (2 hours), and a test set with 1572 samples (2 hours). It comprises around 300 speakers and more than 100k running words with a vocabulary size of around 10k.

\noindent\textbf{VLRF} \cite{fernandez2017towards} is a Spanish database primarily designed to study the feasibility of the VSR task, providing 1 hour of data. It was recorded in controlled laboratory settings, providing a speaker-dependent partition composed of 24 speakers, each one with 25 assigned unrelated but predefined sentences. It is split into a training set with 480 samples (0.92 hours) and a test set with 120 samples (0.23 hours), comprising around 4k running words with a vocabulary size of 1374.

\subsection{Model Architecture Comparison.}
In our work, we compared three different types of models.

\noindent\textbf{Audio- and Video-Only.} These modality-specific systems constitute the baseline from which we will extract the Branchformer-based knowledge to design our proposed tailored AVSR system. Note that in these audio- and video-only settings, the model architecture described in Subsection \ref{sec:architecture} is modified so that, in addition to removing the audio-visual fusion module, only the corresponding frontend is considered. The audio- and video-only models comprise around 51.2M and 60.7M parameters, respectively, for both languages.

\noindent\textbf{Conventional AVSR.} This model corresponds to the architecture described in Subsection \ref{sec:architecture}, where a modality-specific Branchformer encoder is defined for each modality. This model represents the conventional architecture adopted in the context of AVSR, comprising around 103.5M parameters for both languages studied in our work.

\noindent\textbf{Tailored AVSR.} This model corresponds to the architecture described in Subsection \ref{sec:tailored}, where a tailored audio-visual Branchformer was designed. This model represents a step forward in parameter efficiency in the context of audio-visual speech integration, comprising around 59.3M or 58.3M parameters for English and Spanish, respectively.

\subsection{Implementation Details} \label {sec:implementation}

\noindent\textbf{Audio Processing.} We first sampled all audio waveforms at 16 kHz and then applied utterance-level mean subtraction.

\noindent\textbf{Video Processing.} All videos were first sampled at 25 fps. Similar to \cite{ma2021conformers,ma2022visual,burchi2023audio}, we cropped patches of 96$\times$96 pixels centered on the speaker's mouth. We used the RetinaFace face detector \cite{deng2020retina} and the Face Alignment Network \cite{bulat2017facealign} to detect 68 facial landmarks. Subsequently, we applied a similarity transformation w.r.t. a neutral reference frame, thereby removing possible translation and scaling variations. The cropped lip regions were converted into gray-scale, and normalized based on the overall mean and variance of the training set.

\noindent\textbf{Text Processing.} We first normalize texts by punctuation removal and uppercasing. Similar to \cite{ma2022visual}, we then used a character-level tokenizer with a vocab size of 41 and 37 for English and Spanish, respectively. In both cases, special symbols, such as the ‘space’, 'end of sentence', and the ‘blank’ tokens, were included.

\noindent\textbf{Data Augmentation.} We applied SpecAugment \cite{park2020specaugment} on our audio cues during training with two frequency masks with mask size parameter F = 27 and five time masks with adaptive size pS = 0.05. Furthermore, we introduce speed perturbation by randomly selecting a scale factor between 0.9 and 1.1 from a uniform distribution. Regarding video, random cropping of 88$\times$88 pixels, horizontal flipping, and time masking \cite{ma2022visual} with a maximum of 0.4 seconds were applied during training. 

\noindent\textbf{Training Settings.} Similar to \cite{burchi2023audio}, we used the Adam optimizer \cite{kingma2014adam} and a Noam scheduler \cite{vaswani2017attention} with 10k warmup steps, yielding a peak learning rate of 0.001 with a batch size of 64 samples. All the proposed models were trained for 100 epochs. Those utterances that were longer than 20 seconds were discarded. In all cases, the loss weight $\alpha$ in Eq. (\ref{eq:ctc-attention-loss}) was set to 0.1. Similar to most works in the field, the visual frontend was initialized in all our experiments with the publicly released weights\footnote{\url{https://github.com/mpc001/Lipreading_using_Temporal_Convolutional_Networks}} trained on the word-level LRW classification task \cite{ma2021towards}. Additionally, encountering convergence issues with the Spanish video-only system and inspired by the work carried out by Ma et al. \cite{ma2022visual}, we adopted a cross-lingual learning strategy, initializing both the visual frontend and the encoder backbone with the weights estimated for the English benchmark. Subsequently, we conducted fine-tuning following the training settings outlined in \cite{ma2022visual}.

\noindent\textbf{Inference Settings.} Similar to \cite{ma2022visual}, we set the LM weight $\beta$ of Eq. (\ref{eq:decoding}) to 0.6 and 0.4 for English and Spanish, respectively. For English, we set a word insertion penalty of 0.5 and a beam size of 40, while for Spanish, we set the word insertion penalty and the beam size to 0.0 and 30, respectively. Regarding the CTC/Attention balance, we set the weight $\lambda$ of Eq. (\ref{eq:decoding}) to 0.1. Note that, during inference, we averaged the model parameters on the 10 epochs with the top validation performance.

\noindent\textbf{Evaluation Metric.} Reported results were evaluated by the well-known Word Error Rate (WER) with 95\% confidence intervals using the bootstrap method described by \cite{bisani2004bootstrap}.

\noindent\textbf{Limitations.} Numerous studies \cite{afouras2018deep,ma2021conformers,shi2022learning,burchi2023audio} investigated the benefits of techniques involving either masking or incorporating noise into acoustic cues during training. While these methods have demonstrated improvements in the robustness of AVSR systems, we consciously refrained from adopting these strategies. Our rationale was to properly study the cross-modality learning capabilities and noise robustness of our proposed method strictly from an architectural design perspective, without incorporating such data augmentation artifacts.

\section{Results \& Discussion}
\label{sec:results}

\begin{table*}[!ht]
\caption{Comparison of WER (\%) to the state of the art on the test sets of the datasets composing the English benchmark for audio-only (A), video-only (V), and audio-visual (AV) settings. Works using non-publicly resources were not considered. \colorbox{gray!20}{Gray shading} highlights, along with their corresponding results, those works comparable to ours in terms of either the number of model parameters, hours of training data, or both. Within each subdivision, best results for each dataset are indicated in \textbf{bold}.}
\label{tab:english}
\scriptsize
\centering
\begin{adjustbox}{max width=\textwidth}
\begin{tabular}{lcccccc}
 \toprule
 \multirow{2}{*}[0pt]{\textbf{Method}} & \multirow{2}{*}[-2pt]{\textbf{\begin{tabular}[c]{@{}c@{}}Modality\end{tabular}}} & \multirow{2}{*}[-2pt]{\textbf{\begin{tabular}[c]{@{}c@{}}No. of\\Parameters\end{tabular}}} & \multirow{2}{*}[-2pt]{\textbf{\begin{tabular}[c]{@{}c@{}}Total\\Hours\end{tabular}}} & \multirow{2}{*}[-2pt]{\textbf{\begin{tabular}[c]{@{}c@{}}Additional\\Datasets\end{tabular}}} & \multicolumn{2}{c}{\textbf{Dataset}} \\ \cmidrule{6-7}
 & & & & & \textbf{LRS2-BBC} & \textbf{LRS3-TED} \\ \midrule

 Shi et al. \cite{shi2022learning} & A & 325M & 2,192 & VoxCeleb2 & - & 1.5\\
 Ma et al. \cite{ma2023auto} & A & 250.4M & 3,448 & LRW, VoxCeleb2, AVSpeech & \textbf{1.5} & 1.0\\
 Burchi and Timofte \cite{burchi2024ctcrnnt} & A & 119M & 3,116 & LRW, VoxCeleb2, AVSpeech & - & \textbf{0.7}\\ \hdashline\noalign{\vskip 0.5ex}
 \colorbox{gray!20}{Ma et al. \cite{ma2021conformers}} & A & $\sim$100M & \colorbox{gray!20}{381 / 595} & LRW & \colorbox{gray!20}{3.9} & \colorbox{gray!20}{2.3}\\
 \colorbox{gray!20}{Ma et al. \cite{ma2023auto}} & A & 250.4M &\colorbox{gray!20}{818} & LRW & \colorbox{gray!20}{2.6} & \colorbox{gray!20}{\textbf{1.5}}\\
 \colorbox{gray!20}{Burchi and Timofte \cite{burchi2023audio}} & A & \colorbox{gray!20}{61.7M} & \colorbox{gray!20}{818} & LRW & \colorbox{gray!20}{\textbf{2.4}} & \colorbox{gray!20}{2.0}\\ \midrule  
 Ours (Conventional) & A & 51.2M & 818 & LRW & \textbf{3.9{\tiny$\pm$0.5}} & \textbf{2.4{\tiny$\pm$0.4}} \\
 Ours (Tailored) & A & 43.3M & 818 & LRW & 4.1{\tiny$\pm$0.5} & 2.6{\tiny$\pm$0.4} \\ \midrule \midrule

 Shi et al. \cite{shi2022learning} & V & 325M & 2,192 & VoxCeleb2 &  - & 26.9\\
 Liu et al. \cite{liu2023synthvsr} & V & \textgreater250M & 6720 & LRW, VoxCeleb2, AVSpeech, Synthetic & - & \textbf{16.9}\\
 Ma et al. \cite{ma2023auto} & V & 250.4M & 3,448 & LRW, VoxCeleb2, AVSpeech & \textbf{14.6} & 19.1\\
 Burchi and Timofte \cite{burchi2024ctcrnnt} & V & 130M & 3,116 & LRW, VoxCeleb2, AVSpeech & - & 25.5\\ \hdashline\noalign{\vskip 0.5ex}
 \colorbox{gray!20}{Ma et al. \cite{ma2021conformers}} & V & $\sim$100M & \colorbox{gray!20}{381 / 595} & LRW & \colorbox{gray!20}{37.9} & \colorbox{gray!20}{43.3}\\
 \colorbox{gray!20}{Prajwal et al. \cite{prajwal2022sub}} & V & - & \colorbox{gray!20}{698} & - & \colorbox{gray!20}{28.9} &\colorbox{gray!20}{40.6}\\
\colorbox{gray!20}{Ma et al. \cite{ma2022visual}} & V & \colorbox{gray!20}{52M} & \colorbox{gray!20}{818} & LRW & \colorbox{gray!20}{\textbf{27.3}} & \colorbox{gray!20}{34.5}\\
 \colorbox{gray!20}{Ma et al. \cite{ma2023auto}} & V & 250.4M & \colorbox{gray!20}{818} & LRW & \colorbox{gray!20}{27.9} & \colorbox{gray!20}{\textbf{33.0}}\\
 \colorbox{gray!20}{Burchi and Timofte \cite{burchi2023audio}} & V & \colorbox{gray!20}{61.7M} & \colorbox{gray!20}{818} & LRW & \colorbox{gray!20}{29.8} & \colorbox{gray!20}{37.5}\\ \midrule 
 Ours (Conventional) & V & 60.7M & 818 & LRW & \textbf{29.9{\tiny$\pm$1.6}} & \textbf{37.0{\tiny$\pm$1.6}}\\
 Ours (Tailored) & V & 51.3M & 818 & LRW & 33.6{\tiny $\pm$1.7} & 41.2{\tiny $\pm $1.7}\\ \midrule \midrule

 Shi et al. \cite{shi2022learning} & AV & 325M & 2,192 & VoxCeleb2 & - & 1.3\\
 Ma et al. \cite{ma2023auto} & AV & 250.4M & 3,448 & LRW, VoxCeleb2, AVSpeech & \textbf{1.5} & 0.9\\
 Burchi and Timofte \cite{burchi2024ctcrnnt} & AV & 197M & 3,116 & LRW, VoxCeleb2, AVSpeech & - & \textbf{0.8}\\ \hdashline\noalign{\vskip 0.5ex}
 \colorbox{gray!20}{Ma et al. \cite{ma2021conformers}} & AV & \colorbox{gray!20}{$\sim$100M} & \colorbox{gray!20}{381 / 595} & LRW & \colorbox{gray!20}{\textbf{3.7}} & \colorbox{gray!20}{2.3}\\
 \colorbox{gray!20}{Burchi and Timofte \cite{burchi2023audio}} & AV & \colorbox{gray!20}{61.7M} & \colorbox{gray!20}{818} & LRW & \colorbox{gray!20}{2.3} & \colorbox{gray!20}{\textbf{1.8}}\\ \midrule
 Ours (Conventional) & AV & 103.5M & 818 & LRW & 3.0{\tiny$\pm$0.4} & 2.4{\tiny$\pm$0.4}\\
 Ours (Tailored) & AV & 59.3M & 818 & LRW & \textbf{3.0{\tiny$\pm$0.4}} & \textbf{2.1{\tiny$\pm$0.4}}\\
 \bottomrule
\end{tabular}
\end{adjustbox}
\end{table*}
\begin{table*}[!htbp]
\caption{Comparison of WER (\%) to the state of the art on the Spanish benchmark test sets for audio-only (A), video-only (V), and audio-visual (AV) settings. Works using non-publicly resources were not considered. For each dataset, best results within each modality subdivision are highlighted in \textbf{bold}. Cells with missing number of model parameters correspond to approaches of different nature based on HMMs \cite{lrec2022liprtve,gimeno2024continuous,adriana2022alr}}
\label{tab:spanish}
\scriptsize
\centering
\begin{adjustbox}{max width=\textwidth}
\begin{threeparttable}

\begin{tabular}{lccccc}
 \toprule
 \textbf{Method} & \textbf{Modality} & \textbf{No. of Parameters} & \textbf{Total Hours} & \textbf{Additional Datasets} & \textbf{\%WER} \\ \midrule
 
 & & & & & \\ \midrule\midrule
 \multicolumn{6}{c}{\textbf{MuAViC\textsubscript{es}}} \\ \midrule\midrule
 Anwar et al. \cite{anwar23muavic} & A & 325M & 2,370 & LRW, LRS3, VoxCeleb2 & 16.5\\
 Burchi et al. \cite{burchi2024ctcrnnt} & A & 119M & 4,957 & LRW, LRS3, VoxCeleb2, AVSpeech, MuAViC & \textbf{7.9} \\ \hdashline\noalign{\vskip 0.5ex}
 Ours & A & 51.2M & 203 & - & 11.6{\tiny $\pm$0.8}\\
 Ours (Tailored) & A & 43.3M & 203 & - & 12.8{\tiny $\pm$0.8} \\
 \midrule
 Ma et al. \cite{ma2022visual}\tnote{$^\dagger$} & V & 52M & 1,546 & LRW, LRS2\&3, AVSpeech & 56.6{\tiny$\pm$0.3}\\
 Kim et al. \cite{kim2023lip}\tnote{$^\dagger$} & V & 325M & 2,825 & LRW, LRS3, mTEDx, MLS & 70.2 \\
 Yeo et al. \cite{yeo2024limited} & V & 250.4M & 3,832 & LRW, LRS2\&3, VoxCeleb2, AVSpeech & \textbf{45.7}\\ \hdashline\noalign{\vskip 0.5ex}
 Ours & V & 60.7M & 1,021 & LRW, LRS2\&3 & 77.4{\tiny $\pm$1.5} \\
 Ours (Tailored) & V & 51.3M & 1,021 & LRW, LRS2\&3 & 80.9{\tiny $\pm$1.6}\\
 \midrule
 Anwar et al. \cite{anwar23muavic} & AV & 325M & 2,370 & LRW, LRS3, VoxCeleb2 & 15.9\\
 Hong et al. \cite{hong2023intuitive} & AV & $>$325M & 2,900 & LRW, LRS3, VoxCeleb2, MuAViC & 18.0\\
 Burchi et al. \cite{burchi2024ctcrnnt} & AV & 197M & 4,957 & LRW, LRS3, VoxCeleb2, AVSpeech, MuAViC & \textbf{8.2}\\ \hdashline\noalign{\vskip 0.5ex}
 Ours (Conventional) & AV & 103.5M & 361 & LRW & 11.6{\tiny $\pm$0.7}\\
 Ours (Tailored) & AV & 58.3M & 361 & LRW & 14.3{\tiny $\pm$1.1} \\
 \bottomrule
 
 & & & & & \\ \midrule\midrule
 \multicolumn{6}{c}{\textbf{CMU-MOSEAS\textsubscript{es}}} \\ \midrule\midrule
 Ma et al. \cite{ma2022visual} & A & 50M & 1,546 & LRW, LRS2\&3, AVSpeech & 15.4{\tiny$\pm$0.1}\\ \hdashline\noalign{\vskip 0.5ex}
 Ours & A & 51.2M & 203 & - & \textbf{12.9{\tiny $\pm$1.2}}\\
 Ours (Tailored) & A & 43.3M & 203 & - & 15.1{\tiny $\pm$1.3} \\ \midrule
 Ma et al. \cite{ma2022visual} & V & 52M & 1,546 & LRW, LRS2\&3, AVSpeech & \textbf{44.6{\tiny$\pm$0.6}}\\ \hdashline\noalign{\vskip 0.5ex}
 Ours & V & 60.7M & 1,021 & LRW, LRS2\&3 & 46.7{\tiny $\pm$2.1} \\
 Ours (Tailored) & V & 51.3M & 1,021 & LRW, LRS2\&3 & 52.6{\tiny $\pm$2.2}\\ \midrule
 Ours (Conventional) & AV & 103.5M & 361 & LRW & \textbf{13.0{\tiny $\pm$1.3}}\\
 Ours (Tailored) & AV & 58.3M & 361 & LRW & 15.3{\tiny $\pm$1.4} \\
 \bottomrule
 
 & & & & & \\ \midrule\midrule
 \multicolumn{6}{c}{\textbf{LIP-RTVE}} \\ \midrule\midrule
 Ours (Prior Work) \cite{lrec2022liprtve} & A & - & 9 & - & 15.3{\tiny $\pm$0.8} \\ \hdashline \noalign{\vskip 0.5ex}
 Ours & A & 51.2M & 203 & - & \textbf{7.8{\tiny $\pm$0.5}}\\
 Ours (Tailored) & A & 43.3M & 203 & - & 8.9{\tiny $\pm$0.5} \\ \midrule
 Ours (Prior Work) \cite{lrec2022liprtve} & V & - & 9 & - & 95.9{\tiny $\pm$0.2}\\ \hdashline\noalign{\vskip 0.5ex}
 Ours & V & 60.7M & 1,021 & LRW, LRS2\&3 & \textbf{58.4{\tiny $\pm$1.0}} \\
 Ours (Tailored) & V & 51.3M & 1,021 & LRW, LRS2\&3 & 64.0{\tiny $\pm$1.1}\\ \midrule
 Ours (Conventional) & AV & 103.5M & 361 & LRW & \textbf{6.7{\tiny $\pm$0.4}}\\
 Ours (Tailored) & AV & 58.3M & 361 & LRW & 7.6{\tiny $\pm$0.5} \\
 \bottomrule

 & & & & & \\ \midrule\midrule
 \multicolumn{6}{c}{\textbf{VLRF}} \\ \midrule\midrule
 Ours (Prior Work) \cite{gimeno2024continuous} & A & - & 1 & - & 9.1{\tiny$\pm$3.0}\\ \hdashline\noalign{\vskip 0.5ex}
 Ours & A & 51.2M & 203 & - & \textbf{4.9{\tiny$\pm$2.1}}\\
 Ours (Tailored) & A & 43.3M & 203 & - & 6.0{\tiny $\pm$2.3} \\ \midrule
 Fernández-López \cite{adriana2022alr} & V & - & - & Synthetic & 72.9 \\
 Ours (Prior Work) \cite{gimeno2024continuous} & V & - & 1 & - & 59.7{\tiny $\pm$4.3} \\ \hdashline\noalign{\vskip 0.5ex}
 Ours & V & 60.7M & 1,021 & LRW, LRS2\&3 & \textbf{21.8{\tiny $\pm$3.0}} \\
 Ours (Tailored) & V & 51.3M & 1,021 & LRW, LRS2\&3 & 23.0{\tiny $\pm$3.2}\\ \midrule 
 Ours (Conventional) & AV & 103.5M & 361 & LRW & \textbf{5.0{\tiny $\pm$1.7}}\\
 Ours (Tailored) & AV & 58.3M & 361 & LRW & 5.8{\tiny $\pm$2.2} \\
 \bottomrule

\end{tabular}
\begin{tablenotes}
    \scriptsize
    \item[$\dagger$] Though results were reported for mTEDx \cite{salesky21mtedx}, it corresponds to MuAViC, as noted by Anwar et al. \cite{anwar23muavic}.
\end{tablenotes}
\end{threeparttable}
\end{adjustbox}
\end{table*}

\noindent\textbf{Supporting our Proposed Method.} Several reasons motivated and supported the design of our unified, tailored audio-visual encoder. First of all, we trained audio- and video-only systems to corroborate that acoustic and visual cues require different contextual dependencies when recognizing speech. As Figure \ref{fig:heatmaps} reflects, the Branchformer encoder adopted a notably different architecture for each modality, with one of the branches often dominating in almost all layers. We hypothesized that the branch associated with the lower relevance could be pruned without a substantial performance loss. Experimental results reported in Tables \ref{tab:english} and \ref{tab:spanish} (rows named as ``Ours" and ``Ours (Tailored)") supported this hypothesis: fine-tuning the tailored version of these modality-specific models for ten epochs yielded recognition rates comparable to their complete counterparts, while reducing the number of parameters by 15.5\%. Therefore, we designed the novel tailored audio-visual framework we present in this paper. Albeit the structure of each branch, as Figure \ref{fig:tailored} suggests, aligns with the patterns observed in the modality-specific models, note that we also introduced two parameter-shared feed-forward modules to exploit the cross-modality capabilities of our architecture.

\noindent\textbf{Effectiveness of the Tailored AVSR framework.} The AVSR task is predominantly assessed in English. For this reason, we first present Table \ref{tab:english} to demonstrate the effectiveness of our proposed parameter-efficient AVSR system, which was able to achieve competitive results compared to the current state of the art. While our approach may be outperformed by various works, they are not directly comparable to our method, since their proposed architectures that not only relied on models comprising vast amounts of parameters, but also on additional large-scale datasets for their pre-training. Similarly, Table \ref{tab:spanish} compares our method within the context of the Spanish benchmarks, where we have surpassed the state of the art in most of the datasets by a substantial margin. However, only in the case of MuAViC, our approach exhibits a significant performance drop compared to \cite{burchi2024ctcrnnt} for ASR and AVSR, and \cite{ma2022visual,yeo2024limited} for VSR. This disparity can be attributed to several factors, including the larger number of parameters and total training hours in their proposed models. Additionally, a notable difference between these models and our work, as well as other underperforming studies \cite{anwar23muavic,kim2023lip,hong2023intuitive}, is the use of the AVSpeech dataset \cite{ephrat18avspeech}. The inclusion of this dataset, primarily compiled from TED talks, may have contributed to the performance boost in these state-of-the-art models by providing in-domain data during their pre-training phase.

Our tailored model stands out as the AVSR architecture with the least parameters, capable of achieving state-of-the-art recognition rates in two different languages, even when trained on a moderate scale of data. Furthermore, given the wide range of data conditions covered by the Spanish benchmark, we demonstrated that our proposed audio-visual framework can effectively address AVSR in diverse domains within a single training procedure.

One interesting finding is that, in both languages, tailored video-only systems showed a significant performance drop compared to their audio-only counterparts, suggesting that VSR is a task more dependent on the model complexity. We also tested the robustness of our AVSR systems in scenarios where acoustic cues are absent and found no significant differences between the complete and tailored architectures.

\begin{table}[!htbp]
\caption{Results in WER (\%) of Our Proposed Audio-Only (A) and Audio-Visual (AV) Speech Recognition Systems in Babble Noisy Settings with Multiple SNR Levels. AV\textsubscript{conv} and AV\textsubscript{tail} refer to the Conventional and Tailored AVSR Systems, Respectively.}
\label{tab:noise}
\scriptsize
\centering
\addtolength{\tabcolsep}{-0.4em}
\begin{tabular}{lcccccc}
 \toprule
 \multirow{2}{*}[0pt]{\textbf{Dataset}} & \multirow{2}{*}[0pt]{\textbf{Method}} & \multicolumn{5}{c}{\textbf{SNR levels (dB)}} \\ \cmidrule{3-7}
 & & -5 & 0 & 5 & 10 & 15 \\ \midrule

 \multirow{3}{*}[0pt]{LRS2-BBC} & A & 72.8{\tiny $\pm$1.8} & 17.4{\tiny $\pm$1.5} & 10.1{\tiny $\pm$0.9} & 5.0{\tiny $\pm$0.6} & 4.3{\tiny $\pm$0.6}\\
 & AV\textsubscript{conv} & 49.4{\tiny $\pm$1.9} & 10.8{\tiny $\pm$1.1} & 6.8{\tiny $\pm$0.7} & 4.2{\tiny $\pm$0.5} & 3.6{\tiny $\pm$0.5}\\
 & AV\textsubscript{tail} & 45.8{\tiny $\pm$1.9} & 11.6{\tiny $\pm$1.2} & 6.0{\tiny $\pm$0.6} & 3.9{\tiny $\pm$0.5} & 3.3{\tiny $\pm$0.5}\\ \midrule
 
 \multirow{3}{*}[0pt]{LRS3-TED} & A & 78.9{\tiny $\pm$1.5} & 18.4{\tiny $\pm$1.6} & 9.1{\tiny $\pm$0.7} & 4.1{\tiny $\pm$0.5} & 2.7{\tiny $\pm$0.4}\\
 & AV\textsubscript{conv} & 62.4{\tiny $\pm$1.7} & 14.8{\tiny $\pm$1.4} & 7.1{\tiny $\pm$0.6} & 3.5{\tiny $\pm$0.4} & 2.8{\tiny $\pm$0.4}\\
 & AV\textsubscript{tail} & 60.5{\tiny $\pm$1.7} & 13.8{\tiny $\pm$1.3} & 6.7{\tiny $\pm$0.6} & 3.6{\tiny $\pm$0.4} & 2.5{\tiny $\pm$0.4}\\ \midrule\midrule
 
 \multirow{3}{*}[0pt]{MuAViC\textsubscript{es}} & A & 94.5{\tiny $\pm$1.4} & 34.5{\tiny $\pm$2.2} & 29.3{\tiny $\pm$1.2} & 16.8{\tiny $\pm$0.9} & 13.3{\tiny $\pm$0.8}\\
 & AV\textsubscript{conv} & 95.5{\tiny $\pm$1.4} & 32.9{\tiny $\pm$2.1} & 28.8{\tiny $\pm$1.2} & 16.5{\tiny $\pm$0.8} & 13.1{\tiny $\pm$0.8}\\
 & AV\textsubscript{tail} & 92.6{\tiny $\pm$1.2} & 34.3{\tiny $\pm$2.1} & 33.3{\tiny $\pm$1.6} & 20.6{\tiny $\pm$1.2} & 16.3{\tiny $\pm$1.1}\\ \midrule
  
 \multirow{3}{*}[0pt]{CMU-MOSEAS\textsubscript{es}} & A & $>$100.0 & 38.2{\tiny $\pm$3.9} & 34.9{\tiny $\pm$2.2} & 19.9{\tiny $\pm$1.5} & 15.5{\tiny $\pm$1.4}\\
 & AV\textsubscript{conv} & $>$100.0 & 38.4{\tiny $\pm$3.5} & 35.2{\tiny $\pm$2.1} & 19.0{\tiny $\pm$1.5} & 14.8{\tiny $\pm$1.3}\\
 & AV\textsubscript{tail} & 98.6{\tiny $\pm$1.5} & 43.5{\tiny $\pm$4.0} & 37.1{\tiny $\pm$2.2} & 23.3{\tiny $\pm$1.7} & 17.6{\tiny $\pm$1.4}\\ \midrule
  
 \multirow{3}{*}[0pt]{LIP-RTVE} & A & 89.6{\tiny $\pm$1.3} & 25.6{\tiny $\pm$1.5} & 18.6{\tiny $\pm$0.8} & 11.0{\tiny $\pm$0.6} & 8.8{\tiny $\pm$0.5}\\
 & AV\textsubscript{conv} & 90.3{\tiny $\pm$1.3} & 25.6{\tiny $\pm$1.6} & 16.3{\tiny $\pm$0.8} & 9.3{\tiny $\pm$0.6} & 7.5{\tiny $\pm$0.5} \\
 & AV\textsubscript{tail} & 87.4{\tiny $\pm$1.0} & 26.9{\tiny $\pm$1.6} & 18.2{\tiny $\pm$0.8} & 10.8{\tiny $\pm$0.6} & 8.6{\tiny $\pm$0.5} \\ \midrule
  
 \multirow{3}{*}[0pt]{VLRF} & A & $>$100.0 & 57.0{\tiny $\pm$14.2} & 34.5{\tiny $\pm$6.0} & 12.8{\tiny $\pm$3.5} & 7.3{\tiny $\pm$2.3}\\
 & AV\textsubscript{conv} & $>$100.0 & 82.4{\tiny $\pm$15.9} & 45.8{\tiny $\pm$8.0} & 12.6{\tiny $\pm$3.0} & 6.2{\tiny $\pm$2.1}\\
 & AV\textsubscript{tail} & $>$100.0 & 59.3{\tiny $\pm$13.1} & 35.4{\tiny $\pm$6.4} & 12.1{\tiny $\pm$3.0} & 6.6{\tiny $\pm$2.5}\\

 \bottomrule
\end{tabular}

\end{table}

\noindent\textbf{Noise Robustness.} As Table \ref{tab:noise} reflects, we evaluated the robustness of our proposed audio-only and audio-visual systems in multiple Signal to Noise Ratio (SNR) levels using babble noise from the NoiseX corpus \cite{varga1993noisex}. Consistent with numerous studies \cite{afouras2018deep,ma2021conformers,burchi2023audio,anwar23muavic}, our experiments show that incorporating visual cues significantly enhances the model performance in high-level noise conditions across all studied benchmarks. Our results demonstrate that, despite a reduction in the number of model parameters by nearly half, our proposed tailored audio-visual encoder retained its robustness to noise. For Spanish, the benefit of incorporating visual cues, as can also be inferred from Table \ref{tab:spanish}, is less pronounced than for English datasets. As we later discuss, this could be attributed to the smaller amount of available visual data, which may limit the robust estimation of the visual encoder.

\begin{figure*}[!htbp]
\centering
\includegraphics[width=0.9\textwidth]{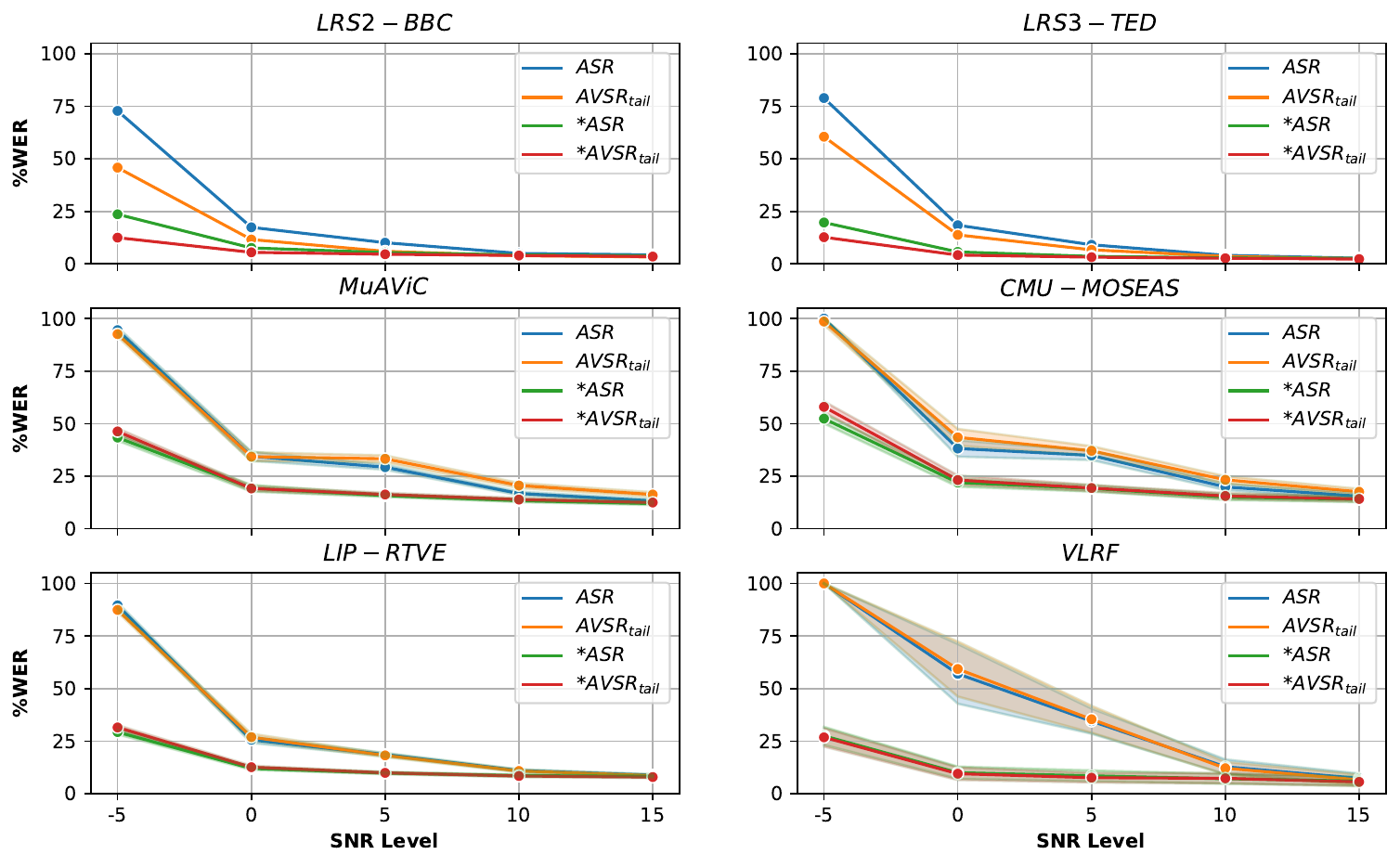}
\caption[Training with Additive Noise on Spanish Bechnmark]{Analysis of training with additive noise both for our proposed audio-only ASR and tailored AVSR architectures. Results in WER (\%) under babble noisy conditions with multiple SNR levels across the English and Spanish test benchmarks. Shaded areas correspond to 95\% confidence intervals. An asterisk (*) indicates experiments where training included babble noise acoustic distortions at varying and random SNR levels.}
\label{fig:addingnoise}
\end{figure*}
\begin{figure*}[h]
\centering
\subfloat[]{\includegraphics[width=2.85in]{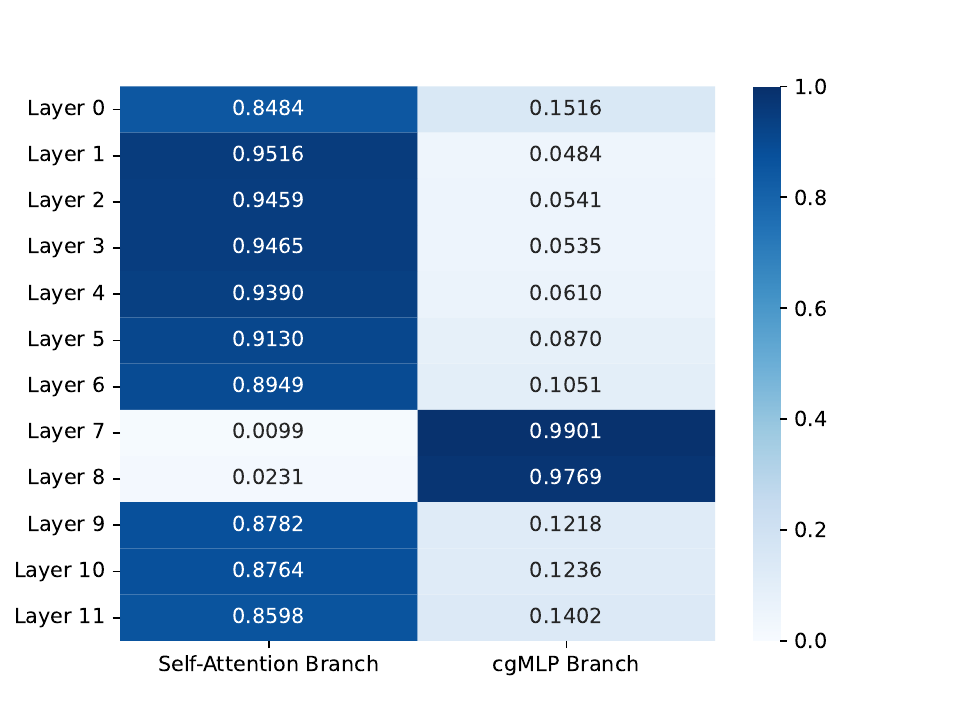}%
\label{asr-heatmap-sp}}
\hfil
\subfloat[]{\includegraphics[width=2.85in]{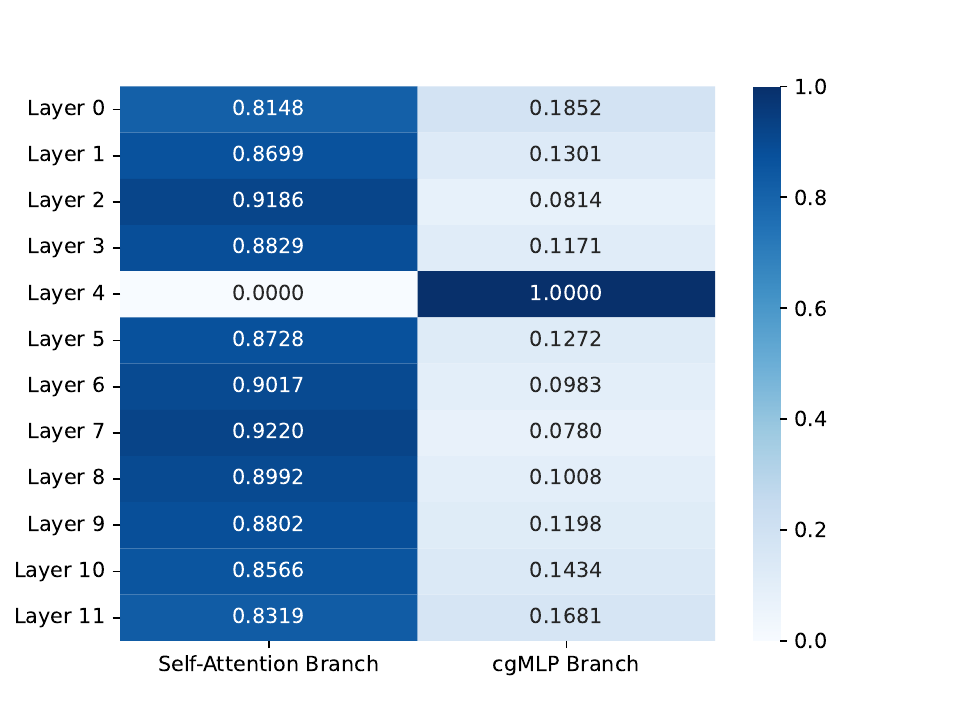}%
\label{vsr-heatmap-sp}}
\caption{Visualization of the average Branchformer weights on the MuAViC validation data set. (a) Acoustic Speech Recognition. (b) Visual Speech Recognition.}
\label{fig:heatmaps_sp}
\end{figure*}

Because of this, additional experiments incorporating noise distortions into acoustic cues during training were also conducted. These data augmentation techniques were expected to increase system robustness to noise and better balance the contributions of audio and visual modalities. As shown in Figure \ref{fig:addingnoise}, while performance under high levels of noise improved significantly, there was no noticeable increase in reliance on visual speech cues, as also evidenced by the poor results when audio input was masked. This suggests that when noise is added during training, the model primarily learns to filter out the noise rather than fully relying on the complementary, yet limited, visual modality. This finding is supported by Lin and Harte \cite{lin2024uncovering}, who argue that current AVSR systems do not fully exploit the visual modality, highlighting the need for further research in this area. However, for English datasets, the visual contribution appears to be more relevant, as previously noted in the discussion of Table \ref{tab:noise}. A potential explanation, particularly regarding discrepancies with English datasets, is the convergence issues observed in video-only Spanish experiments. In these cases, the model failed to achieve satisfactory results unless its encoder was pre-trained with weights from the English model. However, within our proposed framework, this approach is not always practical, as it requires the same tailored architecture across both languages. This further emphasizes the necessity for future work on cross-lingual learning in lipreading technologies.

Interestingly, we also observed differences in how noise distortions were incorporated during training for Spanish and English datasets. While for Spanish, the model was able to handle heavily distorted audio signals (up to -5 SNR) from the beginning, for English, a gradual introduction of noise through curriculum learning was necessary, progressively worsening the SNR level across epochs. This discrepancy likely stems from the cleaner nature of the English data compared to the Spanish corpus, making the incorporation of noise during training more impactful for the former. These observations further support the idea that dataset characteristics play a crucial role in determining model effectiveness in adverse scenarios, such as noisy environments, and even in the modeling of visual cues, as discussed in \cite{gimeno2025evaluation}.

\noindent\textbf{Interpreting AVSR.} In Figure \ref{fig:heatmaps}, we can contrast the different patterns observed between the branch weights of an audio-only and a video-only system. Consistent with the findings of the original Branchformer authors \cite{peng2022branchformer}, the audio-only system adopted an architecture where the relevance of each branch changes primarily in an interleaving manner across layers. Conversely, the video-only system exhibits a high dependency on global-context relationships to effectively process the visual speech features. The model may rely on context due to limited information and lipreading's inherent ambiguity.

As suggested by Figure~\ref{fig:heatmaps_sp}, similar patterns were observed across all benchmarks in our study for video-only systems. This may be largely due to the Spanish VSR models’ strong reliance on English pre-training, which likely guided the model toward similar architectural patterns. In contrast, the audio-only Spanish model --- achieving strong performance without English pre-training --- did not exhibit the same interleaved layer-wise pattern seen in the English model. This discrepancy across languages in terms of model architecture suggests potential differences in how speech information is processed, which may be particularly relevant for the development of multilingual ASR systems~\cite{anwar23muavic,hong2023intuitive}, opening up promising avenues for future research. Another possible reason for this difference is that the Spanish audio-only model was trained on four distinct datasets from different domains, which may have encouraged the encoder to learn more generalizable and context-aware representations.

Regarding our AVSR systems, we also inspected the relative weight learned during training when fusing the audio and visual cues. In all cases, regardless of the language or the architecture, our adaptive fusion module consistently assigned roughly 73.1{\footnotesize $\pm$12.3}\% relevance to the acoustic cues. This observation aligns with Duchnowski et al.~\cite{duchnowski2000development}, which supported that only around 30\% of speech information is visible. 

\begin{figure}[!htbp] 
  \centering
  \begin{tikzpicture}[scale=0.8]
    \begin{axis}[xlabel=\textbf{SNR level (dB)},
                 ylabel=\textbf{\% Acoustic Relative Weight},
                 ymajorgrids=true,
                 xtick=data,
                 label style={font=\small},
                 nodes near coords={$\pgfmathprintnumber{\pgfkeysvalueof{/data point/y}}$},
                 every node near coord/.append style={font=\scriptsize},%
                 nodes near coords style={/pgf/number format/.cd,fixed zerofill,precision=1},%
                 nodes near coords align={anchor=west,yshift=-0.78cm, xshift=-0.5cm},%
                 tick label style={font=\footnotesize},
                 xticklabels={-20,-15,-10,-5,0,5,10,15,20}]
 
    \addplot[mark=o, mark options={solid, thick, green!60!black}, black!75, error bars/.cd, y dir=both, y explicit] 
      plot coordinates {
        (1,62.0) +- (2.6,2.6)
        (2,63.7) +- (2.7,2.7)
        (3,67.8) +- (3.0,3.0)
        (4,72.2) +- (2.7,2.7)
        (5,77.4) +- (3.3,3.3)
        (6,77.4) +- (2.3,2.3)
        (7,78.5) +- (2.1,2.1)
        (8,79.0) +- (2.0,2.0)
        (9,79.3) +- (2.0,2.0)
      };
    \end{axis}
  \end{tikzpicture}
\caption{Average relative acoustic weights learned by the proposed Adaptive Audio-Visual Fusion layer at different SNR levels, evaluated on the test sets of the English benchmarks in our case study. The weights analyzed correspond to the tailored AVSR model architecture trained with noise distortions.}
\label{fig:audiovsnoise}
\end{figure}

Additional experiments were carried out to investigate the variability of the learnable relative weights under different levels of noise degradation. In line with the findings reported by Peng et al.~\cite{peng2022branchformer}, our initial analyses revealed that the weights remained remarkably stable across diverse conditions. However, when noise distortions were introduced during training, notable changes emerged. As shown in Figure~\ref{fig:audiovsnoise}, we can observe how visual cues became increasingly influential as the SNR decreased. Interestingly, these changes when fusing modalities were only markedly pronounced in the tailored, parameter-efficient architecture. This suggests that models with lower complexity are less able to learn to denoise corrupted audio and instead tend to compensate for audio degradation by relying more heavily on visual speech cues. Conversely, for the Spanish benchmarks, the adaptive fusion either remained largely stable across noise conditions (in the full-model dense setup) or exhibited a even stronger dependence on the audio modality around 90\% (in the tailored variant), underscoring ongoing challenges in effectively leveraging visual cues in cross-language settings, as discussed throughout this section.

Overall, our analysis provides initial insights into the architectural differences, beyond the frontends, when dealing with speech features from different modalities and how relevant each of them can be when decoding speech.

\section{Conclusions \& Future Work}

In this paper, we introduced a novel framework for the tailored design of unified audio-visual encoders in the context of parameter-efficient AVSR by leveraging the interpretability and flexibility offered by the Branchformer architecture \cite{peng2022branchformer}. Through extensive experiments across multiple languages and data conditions, we showed the effectiveness of our proposed encoder, which was able to achieve competitive results with a moderate amount of training data, while significantly reducing the number of parameters composing the model. Regarding our future work, we intend to focus our efforts towards developing lightweight AVSR systems for real-world applications. This includes exploring the combination of our tailored audio-visual encoder with parameter-sharing techniques, such as the Sim-T architecture \cite{wei2023sim}; the integration of audio-visual cues in early stages through, for instance, the implementation of intermediate auxiliary losses \cite{ma2022visual,lee2021interctc}; and finally the shift to non-autoregressive paradigms, such as the Mask-CTC decoder \cite{higuchi2021maskctc}, capable of achieving state-of-the-art recognition rates in much less inference time. In this context, we further aim to investigate the potential of cross-modal and cross-lingual learning within parameter-efficient AVSR frameworks.


\section*{Acknowledgments}

This work was partially supported by Generalitat Valenciana through Grant CIACIF/2021/295 and Conselleria de Educación, Universidades y Empleo under project LightVED, funded by the PROMETEO 2024 program (CIPROM/2023/17). It was also supported by Grant PID2021-124719OB-I00 under project LLEER, funded by MCIN/AEI and ERDF, EU ``A way of making Europe".



\bibliographystyle{IEEEtran}
\bibliography{main}


 
%



\vspace{-33pt}
\begin{IEEEbiography}[{\includegraphics[width=1in,height=1.25in,clip,keepaspectratio]{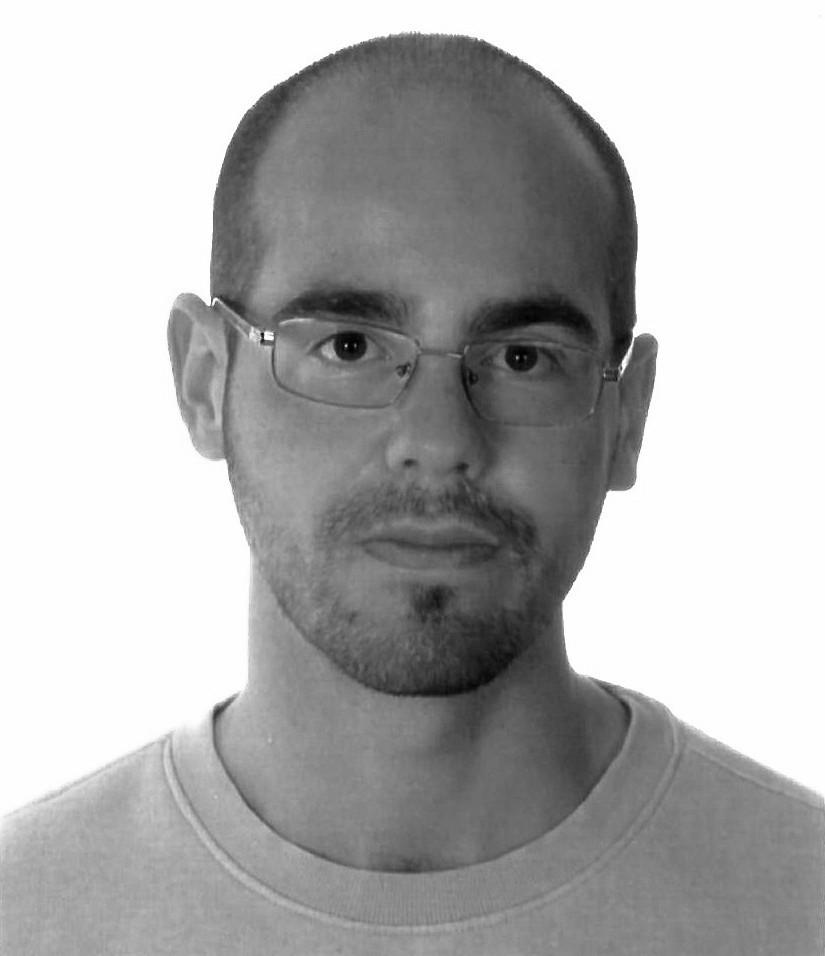}}]{David Gimeno-Gómez} received the B.Sc. degree in computer science and the M.Sc. degree in artificial intelligence, pattern recognition, and digital image from the Universitat Politècnica de València, Valencia, Spain, in 2019 and 2020, respectively. He is currently working towards a thesis for the Ph.D. degree. He pertains to the Pattern Recognition and Human Language Technology Research Center, where he develops his research on the topics of speech recognition, affective computing, and multimodal learning. Since 2020, he has participated in several research projects related to deep learning for multimodal interaction and audio-visual speech recognition.
\end{IEEEbiography}

\vspace{-33pt}
\begin{IEEEbiography}[{\includegraphics[width=1in,height=1.25in,clip,keepaspectratio]{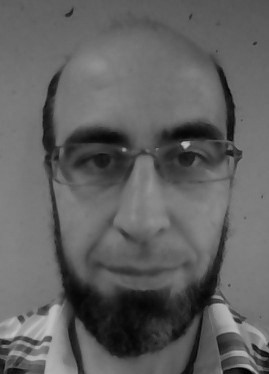}}]{Carlos-D. Martínez-Hinarejos} received the B.Sc. degree in computer science, the Ph.D. degree in pattern recognition and artificial intelligence, and the B.Sc. degree in biotechnology from the Universitat Politècnica de València (UPV), Valencia, Spain, in 1998, 2003, and 2012, respectively. He joined the UPV staff in the Computation and Computer Systems Department, UPV, in 2000. He pertains to the Pattern Recognition and Human Language Technology Research Center, where he develops his research on the topics of speech recognition, dialogue systems, multimodal systems, and text classification. He has participated in many European and Spanish projects, and is the current coordinator of the Spanish Network for Speech Technologies.
\end{IEEEbiography}
\vspace{11pt}


\vfill

\end{document}